\documentclass{article}
\newcommand{\keywords}[1]{%
  \vspace{2ex}
  \noindent\textbf{Keywords:} #1
}
\usepackage{graphicx} 
\usepackage{amsmath,algorithm,algcompatible,amssymb,latexsym,amscd,amsfonts}
\usepackage{microtype}  
\newcommand{\RETURN}{\textbf{Return }}
\DeclareMathAlphabet{\mathpzc}{OT1}{pzc}{m}{it}
\usepackage{upgreek}
\usepackage{graphicx}
\usepackage{float}        
\usepackage{caption}
\usepackage{placeins}     
\usepackage{amsthm}
\usepackage{array,xcolor}
\usepackage{wrapfig,yfonts}
\usepackage{float}
\usepackage{graphicx}
\usepackage[mathscr]{eucal}
\usepackage{mathscinet}
\usepackage{mathtools}
\usepackage{mathptmx}
\usepackage{booktabs}
\usepackage{multirow}
\usepackage[figuresright]{rotating}
\usepackage{lscape}

\numberwithin{equation}{section}

\makeatletter
\renewcommand{\@biblabel}[1]{#1\hfill \hspace{-0.2cm}}
\makeatother
\usepackage{cite}
\title{Data driven feedback linearization of nonlinear control systems via Lie derivatives and stacked regression approach}
 \author{%
    \begin{tabular}{c}
        \textsuperscript{1}Lakshmi Priya P. K. and
        \textsuperscript{2}Andreas Schwung  \\[6pt]
        \multicolumn{1}{c}{\textsuperscript{1,2}Department of Automation Technology and Learning Systems} \\
        \multicolumn{1}{c}{South Westphalia University of Applied Sciences} \\
        \multicolumn{1}{c}{Lübecker Ring 2, Soest, 59494, North Rhine-Westphalia, Germany} \\[6pt]
        \textsuperscript{1}Email: pondicherrykumar.lakshmipriya@fh-swf.de
        \textsuperscript{2}Email: schwung.andreas@fh-swf.de \\
    \end{tabular}%
}

\usepackage[a4paper,margin=1in]{geometry}
\begin{document}
\maketitle
\begin{abstract}
Discovering the governing equations of a physical system and designing an effective feedback controller remains one of the most challenging and intensive areas of ongoing research. This task demands a deep understanding of the system behavior, including the nonlinear factors that influence its dynamics. In this article, we propose a novel methodology for identifying a feedback linearized physical system based on known prior dynamic behavior. Initially, the system is identified using a sparse regression algorithm, subsequently a feedback controller is designed for the discovered system by applying Lie derivatives to the dictionary of output functions to derive an augmented constraint which guarantees that no internal dynamics are observed. Unlike the prior related works, the novel aspect of this article combines the approach of stacked regression algorithm and relative degree conditions to discover and feedback linearize the true governing equations of a physical model. 
\end{abstract}


\section{Introduction}
Inherently, many real-world problems possess nonlinear dynamics by nature, making their behavior complex and often difficult to predict. The modeling and assessment of these systems are more accurately represented through nonlinear control systems, which are specifically designed to handle such complexities. The only unifying factor among nonlinear control systems is the presence of nonlinearities in their dynamics. Due to the lack of generalization, most work has traditionally focused on linearizing the system. However, over time analytical approaches to solve nonlinear systems have been developed \cite{m0,v2, z1}. A few prominent application areas include robotics, aircraft control systems, biomedical engineering devices, and nonlinear electrical circuits \cite{a2, f1,m2}.
  The widespread presence of nonlinearities in these domains has contributed significantly to the development and refinement of theoretical analysis in nonlinear control systems.
  Over the decades, extensive analytical study and control techniques have been developed to support model-based control approaches. Nevertheless, in recent years, it has become increasingly evident that many physical systems lack an accurate or readily available mathematical model. This raises an important question of how to handle such situations. Consequently, there has been a growing emphasis on the study and application of data-driven control systems (DDCS). \\
   \par DDCS are emerging as a viable alternative, leveraging prior knowledge of system behavior and dynamics without requiring a complete analytical model. With the increasing availability of large system datasets, it has become feasible to directly map data to a control law, sometimes even bypassing the need for explicit system equations. In almost every area of nonlinear control, data-driven control techniques have been applied. To mention a few: predictive control, optimal control, robust control, and networked control \cite{w3,b3,v4,m3}. The data-driven technique for designing a control system involves deducing the system equations directly using the data collected from the plant, without requiring prior knowledge of the true underlying dynamics of the system. Willems' Fundamental Lemma demonstrated that it is possible to parametrize the trajectories of a linear system using a sufficiently exciting single trajectory of the same system. This laid a profound foundation for identification of system using some random measured data. Later it was extended to relax the condition stating even multiple short trajectories can be used to obtain all trajectories of a system provided the trajectories are persistently exciting \cite{p3}. \\
\par The initial breakthrough in using training datasets to identify the true dynamics of a governing system was achieved by Hod Lipson \cite{s1}, the study addressed the long-standing challenge of automating the discovery of physical laws from available system data. Symbolic regression is used as an analytical tool to derive the governing nonlinear system to reduce the model's complexity with exact accuracy, and the effectiveness was demonstrated on motion tracking data of a double pendulum. However, a common issue associated with symbolic regression is its tendency to overfit the data. In 2015, a slightly improvised technique was introduced to identify the system using sparse regression \cite{b1}. The underlying idea is based on the assumption that only a few key terms are needed to accurately model a dynamical system. Furthermore, sparse regression combined with machine learning approaches was used to extract the true dynamics of a system from noisy datasets, ultimately resolving a 30-year long-standing problem in fluid dynamics \cite{m2}. This eventually derived the attention for data driven systems resulting in a wide range of advanced techniques \cite{g1}. The crucial development of sparse identification of a system is that it can be used to identify the governing partial differential equations of a physical system with exact accuracy \cite{r1}. For models with large scale training datasets, system identification is often more complex and less accurate due to overfitting issues. This limits the use of data-driven control techniques in neural network systems, as they typically require large volumes of training data. In \cite{k1}, E. Kaiser identified that this problem can be overcome by using sparse regression methods to determine the true equations of dynamical systems. \\
\par The conceptual approach of simplifying the dynamics of a system by representing it in a different state space is known as feedback linearization. This can be broadly viewed as the transformation of a complex system into its simplest equivalent form. Feedback linearization is further classified into direct, partial and exact form. This technique has been used successfully to solve some of the most intricate problems in control theory. Therefore, designing a feedback linearizing controller for data driven system is a central focus in modern control theory \cite{p2, y1}. Despite the wide range of applications of feedback linearization in nonlinear control problems, the need for detailed knowledge of system dynamics remains a significant constraint in many real world scenarios. This limitation is largely due to damping effects and parametric uncertainties inherent in physical systems \cite{a1, l1, v3}.
Systems modeled using training datasets often pose a challenge when designing controllers due to the presence of complex nonlinearities. The reason behind this set back is that most of the classical control methods are designed for linear systems. Therefore, canceling the true nonlinear dynamics enhances the ability to design more accurate and reliable feedback controllers.  Michel Guo, developed a semi definite programming method to stabilize the nonlinear dynamics of a system by using approximate nonlinearity cancellation and establishing a stable region of convergence\cite{g3}. Gaussian regression to develop a feedback controller for an event-triggered nonlinear system was studied in \cite{u1}.
Some of the closely associated works with data-driven control systems include \cite{k2}, where the authors develop the procedure for modeling uncertainty in energy distributed systems using physics-based methods. Interpolation method based on kernel techniques was introduced to develop a data driven nonlinear control systems which cancels the nonlinearities to ensure closed-loop stability \cite{h1}. An identification technique based on sparse regression is used to model dc-converters, where the dynamics of the nonlinear terms are directly discovered using a build in dictionary function \cite{h2}.  C. De Persis, in his work \cite{p4}, introduces a method to linearize the controller using learning techniques applied to datasets. This differs from our proposed approach, which focuses on feedback linearization of data-driven systems based entirely on the use of Lie derivatives applied to the data matrix of output dictionary functions. This allows us to derive augmented constraints that enforce the system's relative degree. In contrast, the earlier work does not explicitly define an output function, and feedback linearization is achieved by identifying the null space through solving a set of linear equations constructed from the training data matrices.
\\
\par Since most physical systems cannot be modeled exactly using mathematical equations, developing a feedback controller for data-driven systems remains a challenging task. In particular, designing feedback controllers with no internal dynamics is an open area of research that is not yet well explored. Therefore, in this article, we propose a new technique to identify the true dynamics of a physical system and derive a novel augmented constraint to ensure the system’s relative degree. The approach combines sparse regression method and Lie derivative concept, applied over a built-in dictionary of functions, to design a feedback linearizing controller. 
The rest of the article is organized as follows: Section 2 discusses the generalized data-driven modeling of the nonlinear control system using a sparse regression algorithm. Section 3 introduces the Lie derivative based feedback linearization approach for data-driven systems. Section 4 formulates the optimization problem along with a bilinear constraint to enforce the relative degree condition required for feedback linearization of a two-state vector system. The results are then extended and generalized to arbitrary systems with no internal dynamics. In Section 5, the proposed method is applied to feedback linearize a two-state Van der Pol oscillator, demonstrating its effectiveness. Finally, Section 6 concludes the discussion.

\section{System identification using sparse regression}
This section re-conceptualizes the problem of discovering the equations of a dynamical control system using the technique of sparse regression. The method solely relies on the fact that almost every physical system has very few terms that actually define the true dynamics. In particular, we address the problem of full state feedback linearization using the approach of Lie derivatives to deduce augmented constraint to establish the relative degree of the system.

Consider a continuous time nonlinear control system of the form
\begin{align}
\dot{\mathrm{x}}(\mathpzc{t}) &= \mathpzc{f}(\mathrm{x}(\mathpzc{t})) + \mathpzc{g}(\mathrm{x}(\mathpzc{t}))\mathpzc{u}(\mathpzc{t}), \nonumber \\
\mathpzc{y}(\mathpzc{t}) &= c(\mathrm{x}(\mathpzc{t})). 
\end{align}
Here, $\mathrm{x}(\mathpzc{t})=[\mathrm{x}_{1}(\mathpzc{t}),\mathrm{x}_{2}(\mathpzc{t}),\mathrm{x}_{3}(\mathpzc{t}),\cdots,\mathrm{x}_{n}(\mathpzc{t})]^T \in \mathscr{R}^{n}$ is the state vector of dimension $n$, $\mathpzc{u}(\mathpzc{t}) \in \mathscr{R}$ is the control input, $\mathpzc{f} \in \mathscr{R}^{n}, \mathpzc{g} \in \mathscr{R}^{n}$ are unknown functions to be identified and the system output is given by $\mathpzc{y}(\mathpzc{t}) \in \mathscr{R}^{n} $. Now to identify the true dynamical equations of the given system, we begin by observing the evolution of the state vector $\mathrm{x}(\mathpzc{t})$ and its derivative $\dot{\mathrm{x}}(\mathpzc{t}).$ Then we have a measurement of data sets sampled over time instances $\mathpzc{t}_{1},\mathpzc{t}_{2},\mathpzc{t}_{3},\cdots,\mathpzc{t}_{m}.$ \\
The matrix representation of the collected data sets are given as :
\begin{equation}
\mathrm{X} =
\begin{bmatrix}
  \mathrm{x}^{T}(\mathpzc{t}_{1})\\
\mathrm{x}^{T}(\mathpzc{t}_{2})\\
\vdots \\
\mathrm{x}^{T}(\mathpzc{t}_{m})
\end{bmatrix}
=
\begin{bmatrix}
  \mathrm{x}_{1}(\mathpzc{t}_{1}) &  \mathrm{x}_{2}(\mathpzc{t}_{1})   & \cdots & \mathrm{x}_{n}(\mathpzc{t}_{1}) \\
  \mathrm{x}_{1}(\mathpzc{t}_{2}) &  \mathrm{x}_{2}(\mathpzc{t}_{2})   & \cdots & \mathrm{x}_{n}(\mathpzc{t}_{2})\\
  \vdots & \vdots & \ddots & \vdots \\
  \mathrm{x}_{1}(\mathpzc{t}_{m}) &  \mathrm{x}_{2}(\mathpzc{t}_{m})   & \cdots & \mathrm{x}_{n}(\mathpzc{t}_{m})
\end{bmatrix}.
\end{equation}

\begin{equation}
\dot{\mathrm{X}} =
\begin{bmatrix}
  \dot{\mathrm{x}}^{T}(\mathpzc{t}_{1})\\
\dot{\mathrm{x}}^{T}(\mathpzc{t}_{2})\\
\vdots \\
\dot{\mathrm{x}}^{T}(\mathpzc{t}_{m})
\end{bmatrix}
=
\begin{bmatrix}
  \dot{\mathrm{x}}_{1}(\mathpzc{t}_{1}) &  \dot{\mathrm{x}}_{2}(\mathpzc{t}_{1})   & \cdots & \dot{\mathrm{x}}_{n}(\mathpzc{t}_{1}) \\
  \dot{\mathrm{x}}_{1}(\mathpzc{t}_{2}) &  \dot{\mathrm{x}}_{2}(\mathpzc{t}_{2})   & \cdots & \dot{\mathrm{x}}_{n}(\mathpzc{t}_{2})\\
  \vdots & \vdots & \ddots & \vdots \\
  \dot{\mathrm{x}}_{1}(\mathpzc{t}_{m}) &  \dot{\mathrm{x}}_{2}(\mathpzc{t}_{m})   & \cdots & \dot{\mathrm{x}}_{n}(\mathpzc{t}_{m})
\end{bmatrix}.
\end{equation}
As a next step, we construct individual libraries $\uptheta (\mathrm{X}, \mathpzc{U}) = [\uptheta_{\mathpzc{f}} (\mathrm{X}), \uptheta_{\mathpzc{g}} (\mathrm{X},\mathpzc{u})] $ for the candidate function $\mathpzc{f}$ and for the function $ \mathpzc{g}$ influenced by the control input $\mathpzc{u}(\mathpzc{t}).$
 \begin{align}
\uptheta_{\mathpzc{f}}(\mathrm{X}) &=
\begin{bmatrix}
\mid & \mid & \mid & \mid & \mid & \mid & \mid & \mid & \mid  \\
1 & \mathrm{X} & \mathrm{X}^{Q_2} & \mathrm{X}^{Q_3} & \cdots & \sin(\mathrm{X}) & \cos(\mathrm{X}) & \sin(2\mathrm{X}) & \cos(2\mathrm{X}) & \cdots \\
\mid & \mid & \mid & \mid & \mid & \mid & \mid & \mid & \mid 
\end{bmatrix},
\end{align}
\begin{align}
\uptheta_{\mathpzc{g}}(\mathrm{X},\mathpzc{u})   &=
\begin{bmatrix}
\mid & \mid & \mid & \mid & \mid & \mid & \mid & \mid & \mid  \\
 \mathpzc{u} & \  \mathrm{X} \, \mathpzc{u} & \mathrm{X}^{Q_2} \, \mathpzc{u} &  \mathrm{X}^{Q_3} \, \mathpzc{u} & \cdots & \sin(\mathrm{X}) \, \mathpzc{u} & \cos(\mathrm{X}) \, \mathpzc{u} & \sin(2\mathrm{X}) \, \mathpzc{u} & \cos(2\mathrm{X}) \, \mathpzc{u}& \cdots \\
\mid & \mid & \mid & \mid & \mid & \mid & \mid & \mid & \mid 
\end{bmatrix},
\end{align}
where $\mathrm{X}^{Q_2}, \mathrm{X}^{Q_3}, \cdots, \mathrm{X}^{Q_n}$ denote the polynomials of higher order. Intuitively, the quadratic nonlinearities are incorporated by the function $\mathrm{X}^{Q_2}$, given by

\begin{equation}
\mathrm{X}^{Q_2}=
    \begin{bmatrix}
        \mathrm{x}_{1}^{2}(\mathpzc{t}_{1})&\mathrm{x}_{1}(\mathpzc{t}_{1})\mathrm{x}_{2}(\mathpzc{t}_{1})  & \cdots & \mathrm{x}_{2}^{2}(\mathpzc{t}_{1})  & \mathrm{x}_{2}(\mathpzc{t}_{1})\mathrm{x}_{3}(\mathpzc{t}_{1})&\cdots& \mathrm{x}_{n}^{2}(\mathpzc{t}_{1})  \\ 
        \mathrm{x}_{1}^{2}(\mathpzc{t}_{2})&\mathrm{x}_{1}(\mathpzc{t}_{2})\mathrm{x}_{2}(\mathpzc{t}_{2})  & \cdots & \mathrm{x}_{2}^{2}(\mathpzc{t}_{2})  & \mathrm{x}_{2}(\mathpzc{t}_{2})\mathrm{x}_{3}(\mathpzc{t}_{2})&\cdots& \mathrm{x}_{n}^{2}(\mathpzc{t}_{2}) \\
        \vdots & \vdots & \ddots & \vdots & \vdots &\ddots & \vdots \\
        \mathrm{x}_{1}^{2}(\mathpzc{t}_{m})&\mathrm{x}_{1}(\mathpzc{t}_{m})\mathrm{x}_{2}(\mathpzc{t}_{m})  & \cdots & \mathrm{x}_{2}^{2}(\mathpzc{t}_{m})  & \mathrm{x}_{2}(\mathpzc{t}_{m})\mathrm{x}_{3}(\mathpzc{t}_{m})&\cdots& \mathrm{x}_{n}^{2}(\mathpzc{t}_{m}) 
 \end{bmatrix}. \nonumber
\end{equation}
The dictionary $\uptheta (\mathrm{X}, \mathpzc{U})$ has $m \times p_{\mathrm{x}}$ dimension where $p_{\mathrm{x}}$ denotes the number of candidate representative functions. Similarly $\uptheta_{\mathpzc{g}}(\mathrm{X},\mathpzc{u})$ is of dimension $m \times p_{\mathpzc{u}}.$ Each column of $\uptheta (\mathrm{X}, \mathpzc{U}) = [\uptheta_{\mathpzc{f}} (\mathrm{X}), \uptheta_{\mathpzc{g}} (\mathrm{X},\mathpzc{u})]$  denotes a candidate function for the right hand side of the system (2.1).  There is no restriction on the construction of the libraries for the candidate functions. We have full freedom to design the entries of the library matrix 
in order to capture the underlying nonlinear dynamics of the system based on the observed data. Therefore this dictionary can include trigonometric functions, polynomials, etc. Since the key idea of this approach relies on the observation that only a small subset of nonlinearities are active in each row of the functions $\mathpzc{f}$ and $\mathpzc{g}$ therefore, we formulate a sparse regression problem to identify the coefficients of the corresponding sparse vectors that govern the dynamics of the model.\\
  
Define
$ \upXi= \big[\Tilde{\upXi},\hat{\upXi}\big],$  where $\Tilde{\upXi}= \big(\Tilde{\upxi}_{1}, \Tilde{\upxi}_{2}, \Tilde{\upxi}_{3}, \cdots , \Tilde{\upxi}_{n}\big)  $ and $\hat{\upXi}= \big(\hat{\upxi}_{1}, \hat{\upxi}_{2}, \hat{\upxi}_{3}, \cdots, \hat{\upxi}_{n}\big) $ are the sparse vector coefficients to determine the nonlinearities of
\begin{align}
\dot{\mathrm{X}}&=\uptheta (\mathrm{X}, \mathpzc{U})\upXi, \nonumber \\
    \dot{\mathrm{X}}& =\uptheta_{\mathpzc{f}} (\mathrm{X}) \Tilde{\upXi} + \uptheta_{\mathpzc{g}} (\mathrm{X},\mathpzc{u})\hat{\upXi},
\end{align}
where, the matrix $\Tilde{\upXi}$ is given by:
\begin{equation}
\Tilde{\upXi}=
    \begin{bmatrix}
        \Tilde{\upxi}_{1,1} & \Tilde{\upxi}_{1,2} & \cdots &\Tilde{\upxi}_{1,n}\\
       
        \Tilde{\upxi}_{2,1} & \Tilde{\upxi}_{2,2} & \cdots &\Tilde{\upxi}_{2,n}\\
        \vdots & \vdots & \ddots & \vdots \\
       
        \Tilde{\upxi}_{{p_{\mathrm{x}}},1} & \Tilde{\upxi}_{{p_{\mathrm{x}}},2} & \cdots &\Tilde{\upxi}_{{p_{\mathrm{x}}},n}
    \end{bmatrix},
\end{equation}
similarly the matrix representation of $ \hat{\upXi}$ can be defined.\\

Since we are working with single input single output system, let  $\upPhi(\mathrm{X})$ be an associated dictionary of function for the output $\mathpzc{y}(\mathpzc{t})$ defined as 
\begin{align}
\upPhi(\mathrm{X})=
    \begin{bmatrix}
        \mid & \mid & \mid & \mid  \\
        1 & \mathrm{x}_{k} &  \mathrm{x}_{k}^{2} & \mathrm{x}_{k}^{3} \cdots \\
        \mid & \mid & \mid & \mid 
    \end{bmatrix}_{m \times p_{\mathpzc{y}}},
\end{align}
with
\begin{align}
\mathrm{Y}=\upPhi(\mathrm{X}) \zeta ,
 \end{align}
where $\zeta \in \mathscr{R}^{p_{\mathpzc{y}} \times 1}$ is a sparse column vector in which most entries are zero, encouraging model sparsity. Once $ \upXi$ is determined successfully, then each row of system (2.1) can be modeled in the following manner
\begin{align}
    \mathrm{x}_{l}=\mathpzc{f}_{l}(\mathrm{x}) + \mathpzc{g}_{l}(\mathrm{x})=\uptheta_{\mathpzc{f}}(\mathrm{x}^{T}) \Tilde{\upxi}_{l}+\uptheta_{\mathpzc{g}}(\mathrm{x}^{T},\mathpzc{u}) \hat{\upxi}_{l}.
\end{align}
At this point it is to be noted that $\uptheta_{\mathpzc{f}}(\mathrm{x}^{T}) $ is a symbolic function in vector form containing elements of $\mathrm{x}.$ Hence,
\begin{align}
    \dot{\mathrm{x}}&=\mathpzc{f}(\mathrm{x})+\mathpzc{g}(\mathrm{x})(\mathpzc{u}), \nonumber\\
  \dot{\mathrm{x}}  &=\upXi^{T}\big(\uptheta(\mathrm{x}^{T}, \mathpzc{u})\big)^{T}.
\end{align} 

\section{Feedback linearization of system}
\par The system considered in this work is modeled as a single-input single-output observation system. The primary objective is to critically examine the notion of relative degree by repeatedly differentiating the output $\mathpzc{y}$ until the control input $\mathpzc{u}$ appears explicitly which is a process formalized through the use of Lie derivatives. The study of Lie derivatives plays a pivotal role in input-output linearization, as it rigorously determines the feasibility and structure of the linearized model. By applying Lie derivatives to a dictionary of candidate output functions, we enable both global identification of the underlying nonlinear dynamics and enforce feedback linearization. This is achieved through the construction of sparsely identified vectors, ensuring a minimal yet expressive representation of the system behavior while preserving its structure. Initially we will discuss for the case where the relative degree of the system $r=2$, later the procedure can be generalized.\\ The nonlinear system (2.1) has relative degree $r$, in terms of Lie derivatives if the following conditions hold \cite{i1}:
\begin{enumerate}
    \item $\mathscr{L}_{\mathpzc{g}}\mathscr{L}^{k}_{\mathpzc{f}}c(\mathrm{x})=0$ for all $\mathrm{x}$ and all $k<r-1$.
    \item  $\mathscr{L}_{\mathpzc{g}}\mathscr{L}^{r-1}_{\mathpzc{f}}c(\mathrm{x})\neq 0$.
\end{enumerate}
$\mathpzc{L}_{\mathpzc{f}} c(\mathrm{x})$ and $\mathpzc{L}_{\mathpzc{g}} c(\mathrm{x})$  are termed as the Lie derivatives of the functions $\mathpzc{f}$ and $\mathpzc{g}$ with the representation
\begin{align}
  \mathpzc{L}^{k}_{\mathpzc{f}} c(\mathrm{x})= \mathpzc{L}_{\mathpzc{f}}\big(\mathpzc{L}^{k-1}_{\mathpzc{f}} c(\mathrm{x})=\dfrac{\partial \mathpzc{L}^{k-1}_{\mathpzc{f}} c(\mathrm{x})}{\partial \mathrm{x}} \mathpzc{f}(\mathrm{x})\big).
\end{align}
Then with a transformation $\mathpzc{q}^{-1}(\mathpzc{q}(\mathrm{x}))=\mathrm{x},$ the new states can be defined as:
\begin{align}
    \vartheta=\begin{bmatrix}
        \vartheta_{1}\\
        \vartheta_{2}\\
        \vartheta_{3}\\
        \vdots\\
        \vartheta_{n}
    \end{bmatrix}=
    \begin{bmatrix}
        \mathpzc{c}(\mathrm{x})\\
        \mathpzc{L}_{\mathpzc{f}}  \mathpzc{c}(\mathrm{x})\\
           \mathpzc{L}_{\mathpzc{f}}^{2}  \mathpzc{c}(\mathrm{x})\\
           \vdots\\
            \mathpzc{L}_{\mathpzc{f}}^{n-1}  \mathpzc{c}(\mathrm{x})\\
    \end{bmatrix}=\mathpzc{q}(\mathrm{x}), \nonumber
\end{align}
and the inverse operator of $\mathpzc{q}(\mathrm{x})$ transforms the  nonlinear system (2.1) to:
\begin{align}
    \begin{bmatrix}
        \dot{\vartheta_{1}}\\
        \dot{\vartheta_{2}}\\
          \vdots \\
           \dot{\vartheta}_{n-1}\\
            \dot{\vartheta_{n}}\\
    \end{bmatrix}&=\begin{bmatrix}
        \vartheta_{2}\\
        \vartheta_{3}\\
        \vdots\\
        \vartheta_{n}\\
          \mathpzc{L}_{\mathpzc{f}}^{n}  \mathpzc{c}(\mathrm{x})
    \end{bmatrix}+ \begin{bmatrix}
        0\\
        0\\
        \vdots\\
     \mathpzc{L}_{\mathpzc{g}}    \mathpzc{L}_{\mathpzc{f}}^{n-1}  \mathpzc{c}(\mathrm{x})
    \end{bmatrix}\mathpzc{u}, \nonumber\\
    \mathpzc{y}&=\vartheta_{1}. \nonumber
\end{align}

As we have already stated previously the system of interest considered in this work is a single-input single-output observation model, when we differentiate the output function $\mathpzc{y}(\mathrm{x})=c(\mathrm{x}),$ the partial derivative of the output function with respect to $\mathrm{x}_{k}$ alone remains non-zero while every other contributes to zero, that is
\begin{align}
\dot{\mathpzc{y}}= \dfrac{d c(\mathrm{x})}{d\mathrm{x}}&=\big[\dfrac{\partial c(\mathrm{x})}{\partial \mathrm{x}_{1}}. \dot{\mathrm{x}}_{1}+\dfrac{\partial c(\mathrm{x})}{\partial \mathrm{x}_{2}}.\dot{\mathrm{x}}_{2}+\cdots,\dfrac{\partial c(\mathrm{x})}{\partial \mathrm{x}_{k}}.\dot{\mathrm{x}}_{k} , \cdots\big], \nonumber \\
    &= \big[ 0, 0, \cdots,\dfrac{\partial c(\mathrm{x})}{\partial \mathrm{x}_{k}}.\dot{\mathrm{x} }_{k}, 0,\cdots \big].
\end{align}
On differentiating the dictionary of function for the output function $\mathpzc{y}=c(\mathrm{x}),$ we get 
\begin{align}
    \dfrac{\partial c(\mathrm{x})}{\partial \mathrm{x}}&=\dfrac{\partial (\upPhi(\mathrm{x}))}{\partial \mathrm{x}} \zeta, \nonumber \\ 
    &= \begin{bmatrix}
    \mid & \mid & \mid & \mid & \\
        0 & 1 & 2 \mathrm{x}_{k} & 3 \mathrm{x}^{2}_{k}  & \cdots \\
        \mid & \mid & \mid & \mid & 
    \end{bmatrix} \zeta, \nonumber  \\ 
    & = \begin{bmatrix}
        0 . \zeta_{1} + 1.  \zeta_{2} + 2 \mathrm{x}_{k} . \zeta_{3} + 3 \mathrm{x}^{2}_{k} .  \zeta_{4}  + \cdots\\
        \vdots
    \end{bmatrix}. 
\end{align}
From eq (2.10) it is know that each row of the function $\mathpzc{g}$ has the form
\begin{align}
 \mathpzc{g}_{1}&=\uptheta_{\mathpzc{g}}(\mathrm{X}, \mathpzc{u})\hat{\upxi}_{1}, \nonumber \\
     \mathpzc{g}_{2}&=\uptheta_{\mathpzc{g}}(\mathrm{X}, \mathpzc{u})\hat{\upxi}_{2},\nonumber \\
     \vdots \nonumber \\
     \mathpzc{g}_{k}&=\uptheta_{\mathpzc{g}}(\mathrm{X}, \mathpzc{u})\hat{\upxi}_{k}.
\end{align}
Therefore, the relative degree constraint $\mathscr{L}_{\mathpzc{g}} c (\mathrm{x})=0$ is given by 
\begin{equation}
    \begin{bmatrix}
        \big( 0 . \zeta_{1} + 1.  \zeta_{2} + 2 \mathrm{x}_{k} . \zeta_{3} + 3 \mathrm{x}^{2}_{1} . \zeta_{4}  + \cdots \big)^1 .  \big( \uptheta_{\mathpzc{g}}(\mathrm{X}, \mathpzc{u})\hat{\upxi}_{k} \big)^1 \\
        \big( 0 . \zeta_{k} + 1.  \zeta_{2} + 2 \mathrm{x}_{k} . \zeta_{3} + 3 \mathrm{x}^{2}_{1} . \zeta_{4}  + \cdots \big)^2 .  \big( \uptheta_{\mathpzc{g}}(\mathrm{X}, \mathpzc{u})\hat{\upxi}_{k} \big)^2 \\
        \vdots \\
        \big( 0 . \zeta_{k} + 1.  \zeta_{2} + 2 \mathrm{x}_{k} . \zeta_{3} + 3 \mathrm{x}^{2}_{1} . \zeta_{4}  + \cdots \big)^m .  \big( \uptheta_{\mathpzc{g}}(\mathrm{X}, \mathpzc{u})\hat{\upxi}_{k} \big)^m
    \end{bmatrix}=0.
\end{equation}
This can be further simplified to an equivalent condition 
\begin{align}
\zeta^{T} \bigg[ \bigg( \dfrac{\partial(\upPhi(\mathrm{x}))}{\partial\mathrm{x}}\bigg)^T . \big( \uptheta_{\mathpzc{g}}(\mathrm{X}, \mathpzc{u}\big)\hat{\upxi}_{k}\bigg]=0.
\end{align}
Now solving the bilinear constraint (3.6) provides our required solution for the feedback linearization of the system (2.1)

\section{Optimization problem along with bilinear constraint}
In order to impose the condition (3.6) let us begin by formulating a optimization problem along with a bilinear constraint to be solved. To do this we simplify the Lie derivative constraint as given below.\\
Let
\begin{align}
    \mathpzc{L}(\mathrm{x})& =
\begin{bmatrix}
0 & 1 & 2\,\mathrm{x}_{k} & 3\,\mathrm{x}^{2}_{k} &  \cdots
\end{bmatrix}^T. 
\end{align}
If we evaluate $\mathpzc{L}(\mathrm{x}(\mathpzc{t})) $ at different time instance say $\mathpzc{t}_{1},\mathpzc{t}_{2},\mathpzc{t}_{3},\cdots,\mathpzc{t}_{m},$ we have the following matrix representation:
\begin{align}
\mathpzc{L}(\mathrm{x}(\mathpzc{t})) =
\begin{bmatrix}
0 & 1 & 2\,\mathrm{x}_{k} (\mathpzc{t}_{1})& 3\,\mathrm{x}^{2}_{k}(\mathpzc{t}_{1}) &  \cdots & n \mathrm{x}^{n-1}_{k}(\mathpzc{t}_{1})  \\
0 & 1 & 2\,\mathrm{x}_{k} (\mathpzc{t}_{2})& 3\,\mathrm{x}^{2}_{k}(\mathpzc{t}_{2}) &  \cdots & n \mathrm{x}^{n-1}_{k}(\mathpzc{t}_{2})\\
\vdots& \vdots&\vdots &\vdots&\ddots&\vdots\\
0 & 1 & 2\,\mathrm{x}_{k} (\mathpzc{t}_{m})& 3\,\mathrm{x}^{2}_{k}(\mathpzc{t}_{m}) &  \cdots & n \mathrm{x}^{n-1}_{k}(\mathpzc{t}_{m})
\end{bmatrix}^T
\end{align}
Then the constraint (3.6) can be simplified into:
\begin{align}
\zeta^{T} M(\mathrm{x}) \hat{\upxi}_{k} &= 0,
\end{align}
with
\begin{align}
M(x)=  \mathpzc{L}(\mathrm{x})(\uptheta_{\mathpzc{g}}(\mathrm{X}, \mathpzc{u}\big)). \nonumber
\end{align}
Therefore we end up with a bilinear problem to be solved and the most prominent method that can be used to solve this would be iteration. \\

\subsection{Sparse identification of nonlinear dynamics}
There are only few terms active in the functions $f$ and $g$ which truly contribute to the system dynamics. Hence, we formulate a sparse problem to solve for the coefficients in $\upXi,$ this way the active terms of $\dot{\mathrm{x}}$ are easily identified.\\
\\
Optimize
\begin{align}
\min_{\Tilde{\upXi} \, ,\hat{\upXi}, \,\zeta} \left\| \dot{\mathrm{X}}-\uptheta_ {\mathpzc{f}}(.)\Tilde{\upXi}-\uptheta_ {\mathpzc{g}}(.)\hat{\upXi} \right\|_{2}^{2} +\min\left\| \mathpzc{y}-\upPhi(.) \zeta\right\|_{2}^{2} + \uplambda_{1} \big( \left\|\Tilde{\upXi} \right\|_{1}+\left\|\hat{\upXi}\right\|_{1}\big)+\uplambda_{2}  \left\| \zeta \right\|_{1},
\end{align}
subject to
\begin{align}
\zeta^{T} M(\mathrm{x}) \hat{\upxi}_{k} &= 0,
\end{align}
the first term of the right hand-side of (4.4) gives the measures of the matched system dynamics while the second term identifies the matched output dynamics from the measured data with $\uplambda_{1}(.)$ and $\uplambda_{2}(.)$ as the corresponding sparsity parameters of the state vectors and the output function.
The $  \mathrm{L}_{2}$ norm $\left\|(.)\right\|_{2}$, acts as the objective function to minimize the error between the estimated derivative and the real derivatives, through the method of iterative least square optimization.\\
\par At this point it is note worthy to understand that the optimization problem (4.4) cannot be solved by using conventional solvers as it is  nonlinear and nonconvex. Hence we use the nonlinear solver fmincon in Matlab to solve it. This can be done by stacking the two different minimization into a single minimization problem along with the bilinear constraint. 
\begin{align}
    \mathpzc{A}_{joint}.\,\eta&=\mathpzc{Z}_{joint} \nonumber ,
    \end{align}
where
\begin{align}
 \underbrace{
  \begin{bmatrix}
  \uptheta_ {\mathpzc{f}}(.)&\uptheta_ {\mathpzc{g}}(.)&0&0&0 \\
    0&0&\uptheta_ {\mathpzc{f}}(.)&\uptheta_ {\mathpzc{g}}(.)&0  \\
    0&0&0&0&\upPhi
  \end{bmatrix}
  }_{\mathpzc{A}_{joint}}
  .\,
  \underbrace{
      \begin{bmatrix}
      \Tilde{\upxi}_{1}\\
      \hat{\upxi}_{1}\\
      \Tilde{\upxi}_{2}\\
      \hat{\upxi}_{2}\\
      \zeta
      \end{bmatrix}
      }_{\eta}
      =
       \underbrace{
      \begin{bmatrix}
      \dot{\mathrm{x}}_{1}\\
      \dot{\mathrm{x}}_{2}\\
      \mathpzc{y}
      \end{bmatrix}
      }_{\mathpzc{Z}_{joint}}.
      \end{align}
Let $P= 2\mathpzc{p}_{x}+2\mathpzc{p}_{u}+\mathpzc{p}_{y},$ then $\mathpzc{A}_{joint}$ is a matrix of dimension $\mathscr{R}^{3m\times P} , \eta \in \mathscr{R}^{P \times 1}$ and $\mathpzc{Z}_{joint}$ is of dimension $\mathscr{R}^{3m \times 1}$. Hence from the above equation (4.6) it is understood $\mathpzc{A}_{joint}$ consists of 3 blocks, one each to model the system (2.1) along with their dictionary of functions.
Then the optimization problem (4.4) and (4.5) is reduced to
\begin{align}
    &\min_{\eta}\left\| \mathpzc{A}\eta - \mathpzc{Z}\right\|_{2}^{2}+ \uplambda\left\|\eta\right\|_{1} , \hspace{0.2cm} \uplambda=\min\{\uplambda_{1},\uplambda_{2}\}\nonumber\\
\text{subject to}\nonumber \\
&\zeta^{T} M(\mathrm{x}) \hat{\upxi}_{k} = 0.
\end{align}
\begin{algorithm}[H]
\caption{Joint Sparse Regression with Bilinear Constraint}
\begin{algorithmic}[1]
 \REQUIRE State data $\mathrm{X},$ estimated derivatives $\dot{\mathrm{X}}$, control input $\mathpzc{u},$ time step $dt$, threshold value $\uplambda$.
\begin{enumerate}
\item Solve the minimization problem (4.7) subject to the defined bilinear constraint.
\item  Define the index sets $\tilde{\upxi}_{1},\title{\upxi}_{2},\hat{\upxi}_{1},\hat{\upxi}_{2},\upzeta$
\item  for $i=1,2,\cdots,m$ do
\item $\lvert \eta \rvert< \uplambda \rightarrow$ $small_{ind}$
\item $\eta(small_{ind})=0$
\item for $i=1:n$ do
\item  $big_{ind}\neq small_{ind}(:,i)$
\item Update the stacked matrix $\mathpzc{A}_{joint}$
\item  end for
\item end for
\end{enumerate}
 \RETURN Stacked coefficient $\eta$ together with identified feedback linearized control system.

\end{algorithmic}
\end{algorithm}

\subsection{Generalization for the case relative degree $r=n$}
In section 4.1, the optimization problem models the feedback linearization of the non-linear control system using Lie derivative constraint $\mathscr{L}_{\mathpzc{g}}\mathscr{L}^{k}_{\mathpzc{f}}c(\mathrm{x})=0$ for $k<r-1, r=2.$ This can further be generalized for the case $r=n.$ The system of interest (2.1) and the construction of dictionaries all remain the same and the procedure is analogous to the case $r=2.$\\
\par To prove the Lie derivative condition $\mathscr{L}_{\mathpzc{g}} \mathscr{L}^{n-2}_{\mathpzc{f}}c(\mathrm{x})=0$, we first evaluate $\mathscr{L}^{k}_{\mathpzc{f}}c(\mathrm{x})$ for $k=1,2,\cdots,n-2$.
In order to make it easy for the readers to understand we consider the following notions:
\begin{align}
    N_{1}(\mathrm{x})&=\bigg(\dfrac{\partial\upPhi(\mathrm{x})}{\partial \mathrm{x}}\bigg)^T \uptheta_{\mathpzc{f}}(\mathrm{X}), \hspace{0.2cm} N_{2}(\mathrm{x})= N_{1}^{'}(\mathrm{x})\uptheta_{\mathpzc{f}}(\mathrm{X}),\hspace{0.2cm} N_{3}(\mathrm{x})= N_{2}^{'}(\mathrm{x})\uptheta_{\mathpzc{f}}(.), \cdots,\nonumber \\
    N_{n-2}(\mathrm{x})&= N_{n-3}^{'}(\mathrm{x})\uptheta_{\mathpzc{f}}(\mathrm{X}). \nonumber
\end{align}
Moreover from the set of equations (2.6)-(2.11) we evaluate
\begin{align}
    \mathscr{L}^{1}_{\mathpzc{f}}c(\mathrm{x})=\zeta^{T} \bigg(\dfrac{\partial\upPhi}{\partial\mathrm{x}}\bigg)^{T}\uptheta_{\mathpzc{f}} (\mathrm{X}) \Tilde{\upXi}&= \zeta^{T} N_{1}(\mathrm{x}) \Tilde{\upXi}, \nonumber\\
     \mathscr{L}^{2}_{\mathpzc{f}}c(\mathrm{x})&=\zeta^{T} N_{2}(\mathrm{x}) \Tilde{\upXi},\nonumber\\
     \mathscr{L}^{3}_{\mathpzc{f}}c(\mathrm{x})&=\zeta^{T} N_{3}(\mathrm{x}) \Tilde{\upXi},\nonumber\\
     \vdots \nonumber \\
      \mathscr{L}^{n-2}_{\mathpzc{f}}c(\mathrm{x})&=\zeta^{T}N_{n-2}(\mathrm{x}) \Tilde{\upXi}, \nonumber\\
      \mathscr{L}_{\mathpzc{g}} \mathscr{L}^{n-2}_{\mathpzc{f}}c(\mathrm{x})&=\zeta^{T}N_{n-2}^{'}(\mathrm{x})\Tilde{\upXi} .\uptheta_{\mathpzc{g}} (\mathrm{X},\mathpzc{u}) \hat{\upXi}. \nonumber
\end{align}
Therefore the relative degree constraint for the feedback linearization of (2.1) can be reformulated to
\begin{align}
  \mathscr{L}_{\mathpzc{g}} \mathscr{L}^{n-2}_{\mathpzc{f}}c(\mathrm{x})&=0, \nonumber\\
 \zeta^{T}N_{n-2}^{'}(\mathrm{x})\Tilde{\upXi} .\uptheta_{\mathpzc{g}} (\mathrm{X},\mathpzc{u}) \hat{\upXi}&=0.  
\end{align}
Hence, the true dynamics of the system can be identified by 
\begin{align}
&\min_{\Tilde{\upXi} \, ,\hat{\upXi}, \,\zeta} \left\| \dot{\mathrm{X}}-\uptheta_ {\mathpzc{f}}(.)\Tilde{\upXi}-\uptheta_ {\mathpzc{g}}(.)\hat{\upXi} \right\|_{2}^{2} +\min\left\| \mathpzc{y}-\upPhi(.) \zeta\right\|_{2}^{2} + \uplambda_{1} \big( \left\|\Tilde{\upXi} \right\|_{1}+\left\|\hat{\upXi}\right\|_{1}\big)+\uplambda_{2}  \left\| \zeta \right\|_{1}, \nonumber\\
\text{subject to}\nonumber\\
&\zeta^{T}N_{n-2}^{'}(\mathrm{x})\Tilde{\upXi} .\uptheta_{\mathpzc{g}} (\mathrm{X},\mathpzc{u}) \hat{\upXi}=0.  
\end{align}
\section{Case study}
In this section, we deal with the combined approach of identification and feedback linearization of  two dimensional Van der Pol oscillator using our proposed approach.\\
Consider
\begin{align}
\dot{\mathrm{x}}_{1} &= \mathrm{x}_{2}, \nonumber \\
\dot{\mathrm{x}}_{2} &=2\vartheta \varsigma \mathrm{x}_{2} -2\vartheta \varsigma\mu\mathrm{x}_{1}^{2}\mathrm{x}_{2} -\vartheta^{2}\mathrm{x}_{1} +\mathpzc{u},\nonumber\\
\mathpzc{y} &= \mathrm{x}_{1}.
\end{align}
For the purpose of simplification, without loss of generality we assume the value one for the system parameters $\vartheta, \varsigma, \mu$. Collect all the values of $\mathrm{x}, \mathpzc{u}, \dot{\mathrm{x}}$ at different time steps $dt,$ with a time span of $m=100.$ As a next step in system identification, individual libraries are built for the functions $\mathpzc{f},\, \mathpzc{g}$ and $\mathpzc{y}.$ The optimization problem (4.7) is solved for the considered oscillator system with fixed sparse threshold value $\uplambda=0.05$ using the nonlinear solver fmincon in Matlab. 
\begin{table}[H]
\centering
\begin{tabular}{l|cccccccc}
\textbf{Sparse vector} & 1 & $u$ & $x_1$ & $x_2$ & $x_1^2 x_2$ & $x_2^2 x_1$ & $x_{1}^{2}u$ \\
\hline
$\tilde{\xi}_1$ & 0 & 0 & 0 & 1.00 & 0 & 0 & 0  \\
$\hat{\xi}_1$   & 0 & 0 & 0 & 0 & 0 & 0 & 0  \\
$\tilde{\xi}_2$ & 0 & 0 & -1.001 & 2.00 & -2.00 & 0.001 & 0 \\
$\hat{\xi}_2$   & -0.0014 & 1.001 & 0 & 0 & 0 & 0 & -0.001  \\
$\zeta$         & 0 & 0 & 1.00 & 0 & 0 & 0 & 0   \\
\end{tabular}

\caption{Sparse coefficient vectors after regression}
\label{tab:sparse_vectors}
\end{table}
Discovered equations after applying the threshold value:
\begin{align}
\dot{\mathrm{x}}_{1}&=\mathrm{x}_{2}, \nonumber\\
\dot{\mathrm{x}}_{2}&=-1.001\mathrm{x}_{1}+2 \mathrm{x}_{2}-2\mathrm{x}_{1}^{2}\mathrm{x}_{2}+1.001\mathpzc{u}, \nonumber \\
\mathpzc{y} &= \mathrm{x}_{1} \nonumber.
\end{align}

\begin{figure}[H]
    \centering
    \includegraphics[width=0.75\linewidth]{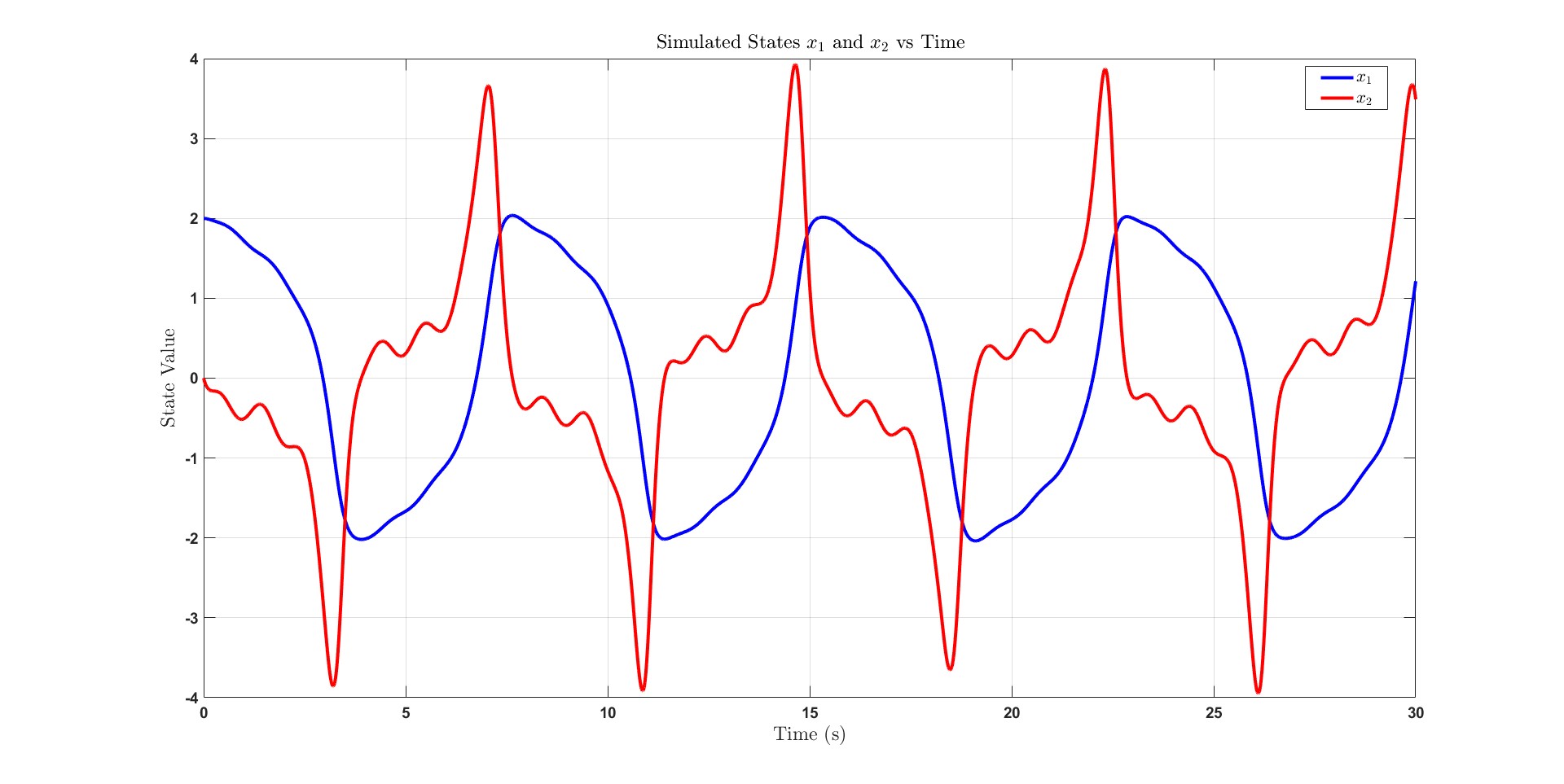}
    \caption{State evolution of the Van der Pol oscillator.}
    \label{fig:states}
\end{figure}

\begin{figure}[H]
    \centering
    \includegraphics[width=1.0\linewidth]{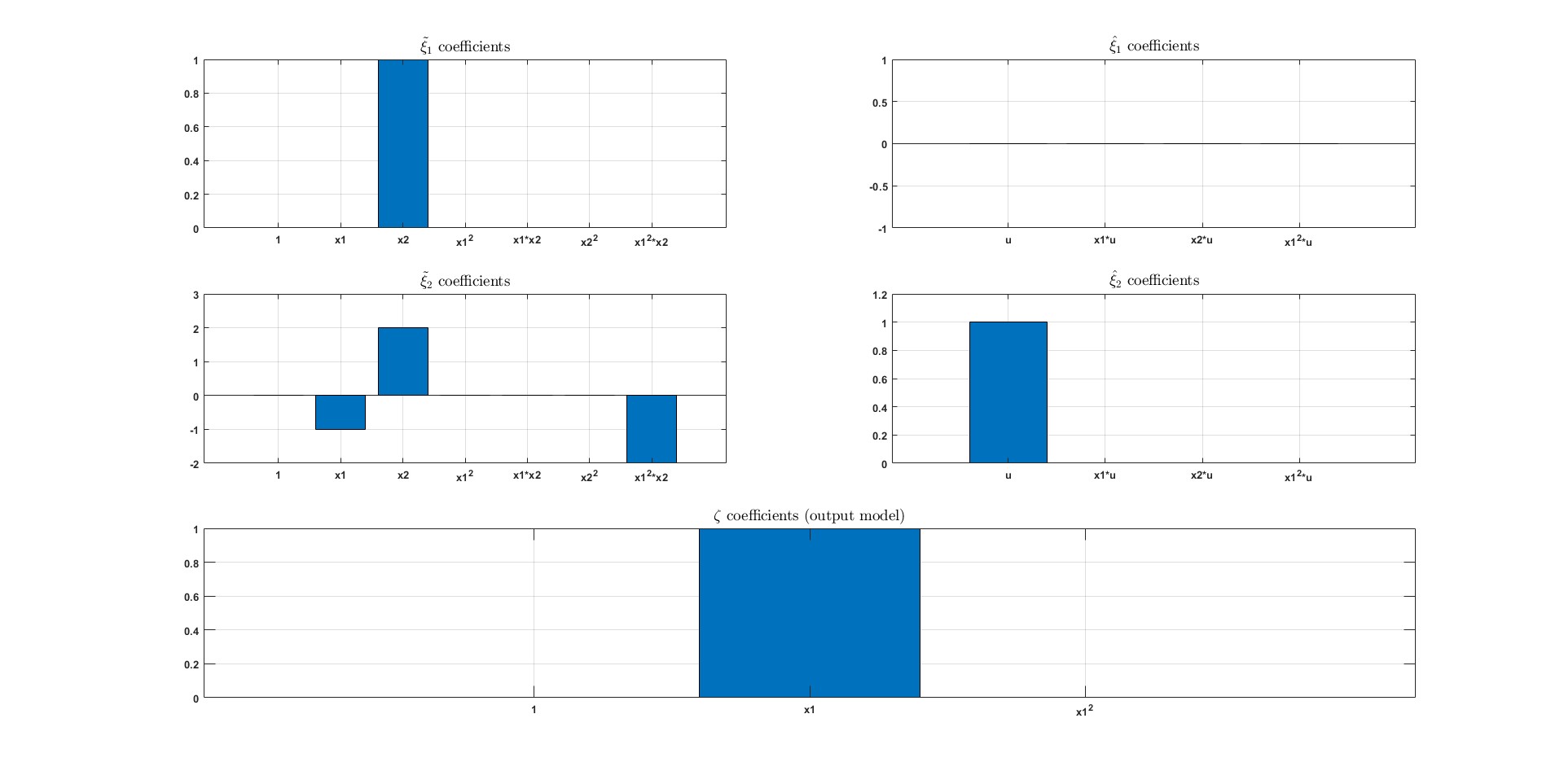}
    \caption{Sparse vector coefficients for the discovered system equations.}
    \label{fig:coefficients}
\end{figure}

\begin{figure}[H]
    \centering
    \includegraphics[width=1.0\linewidth]{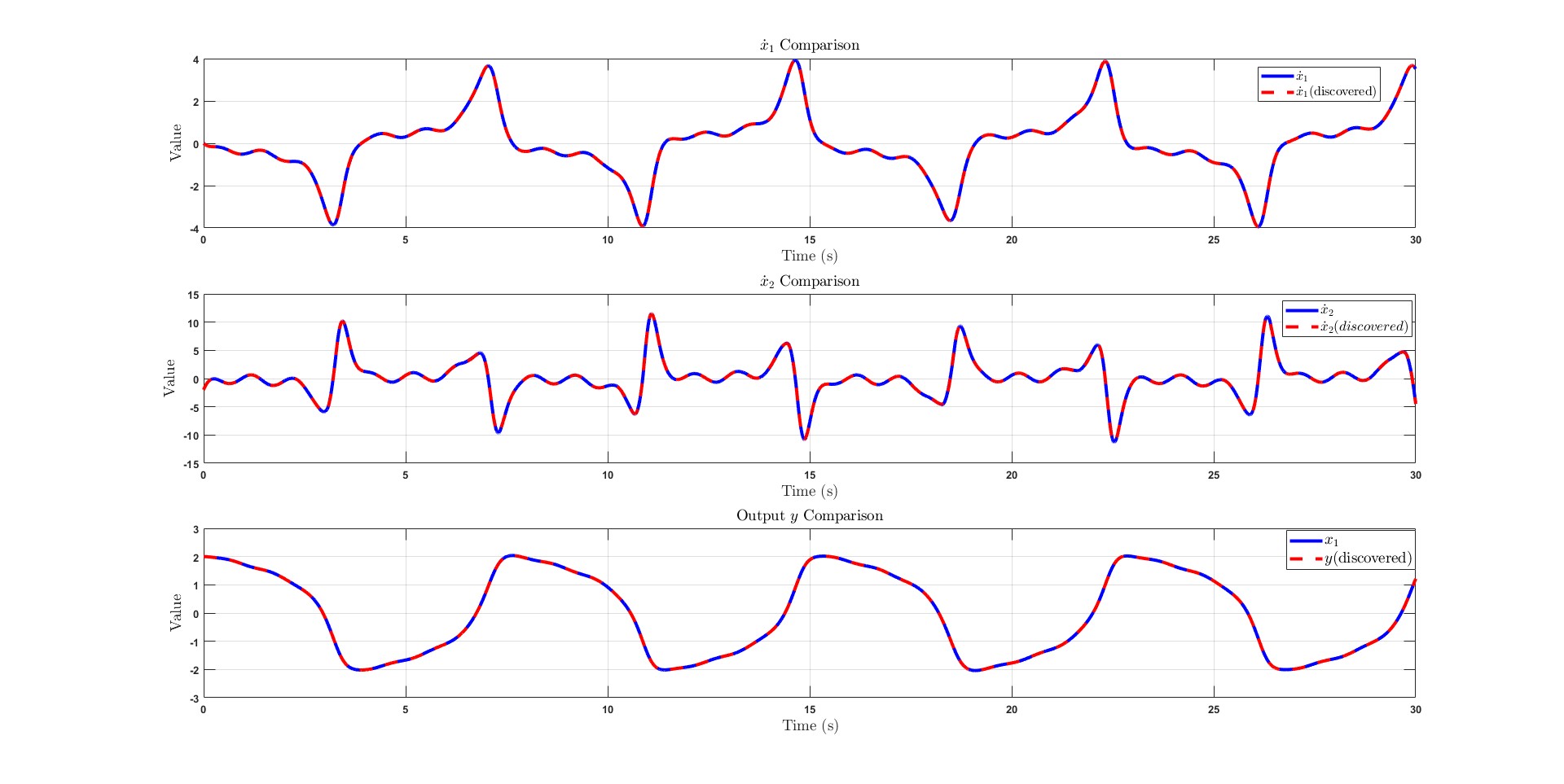}
    \caption{True vs Identified system dynamics.}
    \label{fig:discovered}
\end{figure}
The obtained Lie derivatives are
\begin{align}
 \mathpzc{L}_{\mathpzc{f}}^{0} \mathpzc{c}(\mathrm{x})&=\mathrm{x}_{1}, \nonumber \\
    \mathpzc{L}_{\mathpzc{f}}^{1} \mathpzc{c}(\mathrm{x})&=\begin{bmatrix}
        1&0
    \end{bmatrix}
    \begin{bmatrix}
        \mathpzc{f}_{1}\\
           \mathpzc{f}_{2}\\
    \end{bmatrix}=\mathrm{x}_{2}, \nonumber \\
     \mathpzc{L}_{\mathpzc{f}}^{2} \mathpzc{c}(\mathrm{x})&=\begin{bmatrix}
        0&1
    \end{bmatrix}
    \begin{bmatrix}
        \mathpzc{f}_{1}\\
           \mathpzc{f}_{2}\\
    \end{bmatrix}=-{x}_{1}+2 \mathrm{x}_{2}-2\mathrm{x}_{1}^{2}\mathrm{x}_{2}, \nonumber \\
    \mathpzc{L}_{\mathpzc{g}} \mathpzc{L}_{\mathpzc{f}}^{1} \mathpzc{c}(\mathrm{x})&=\begin{bmatrix}
        0&1
    \end{bmatrix}
   \begin{bmatrix}
        0\\
        1
    \end{bmatrix}
    =1\nonumber
    \end{align}
    On choosing the poles at $s=-2,-6$ the feedback linearized controller is formulated as
    \begin{align}
        \mathpzc{u}=-2\mathrm{x}_{2} +{x}_{1}+2\mathrm{x}_{1}^{2}\mathrm{x}_{2}+5(r-\mathrm{x}_{1})+4(\dot{r}-\mathrm{x}_{2})+\Ddot{r}. \nonumber
    \end{align}
 Two specific cases are analyzed for the data-driven feedback linearization of the Van der Pol oscillator. In the first case, the control input is designed to suppress the inherent oscillatory behavior, effectively stabilizing the system at the origin. In contrast, the second case demonstrates the ability of the proposed control scheme to enable the system to accurately track a desired reference trajectory, highlighting its flexibility and effectiveness in both stabilization and trajectory-tracking scenarios.

\begin{figure}[H]
    \centering
    \includegraphics[width=1.00\linewidth]{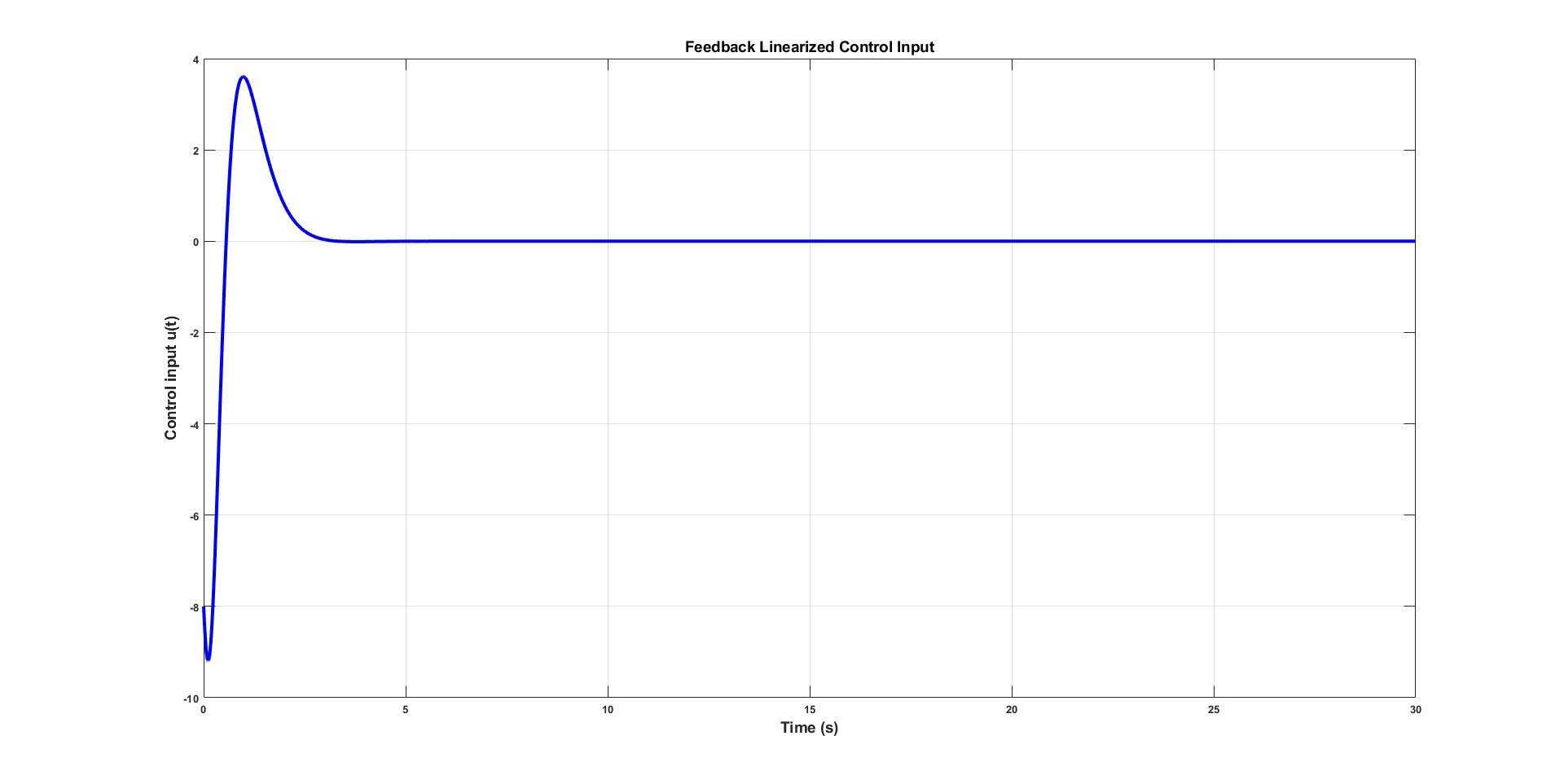}
    \caption{Feedback linearized control input $\mathpzc{u}(\mathpzc{t})$.}
    \label{fig:fdcontrollinear}
\end{figure}

\begin{figure}[H]
    \centering
    \includegraphics[width=1.00\linewidth]{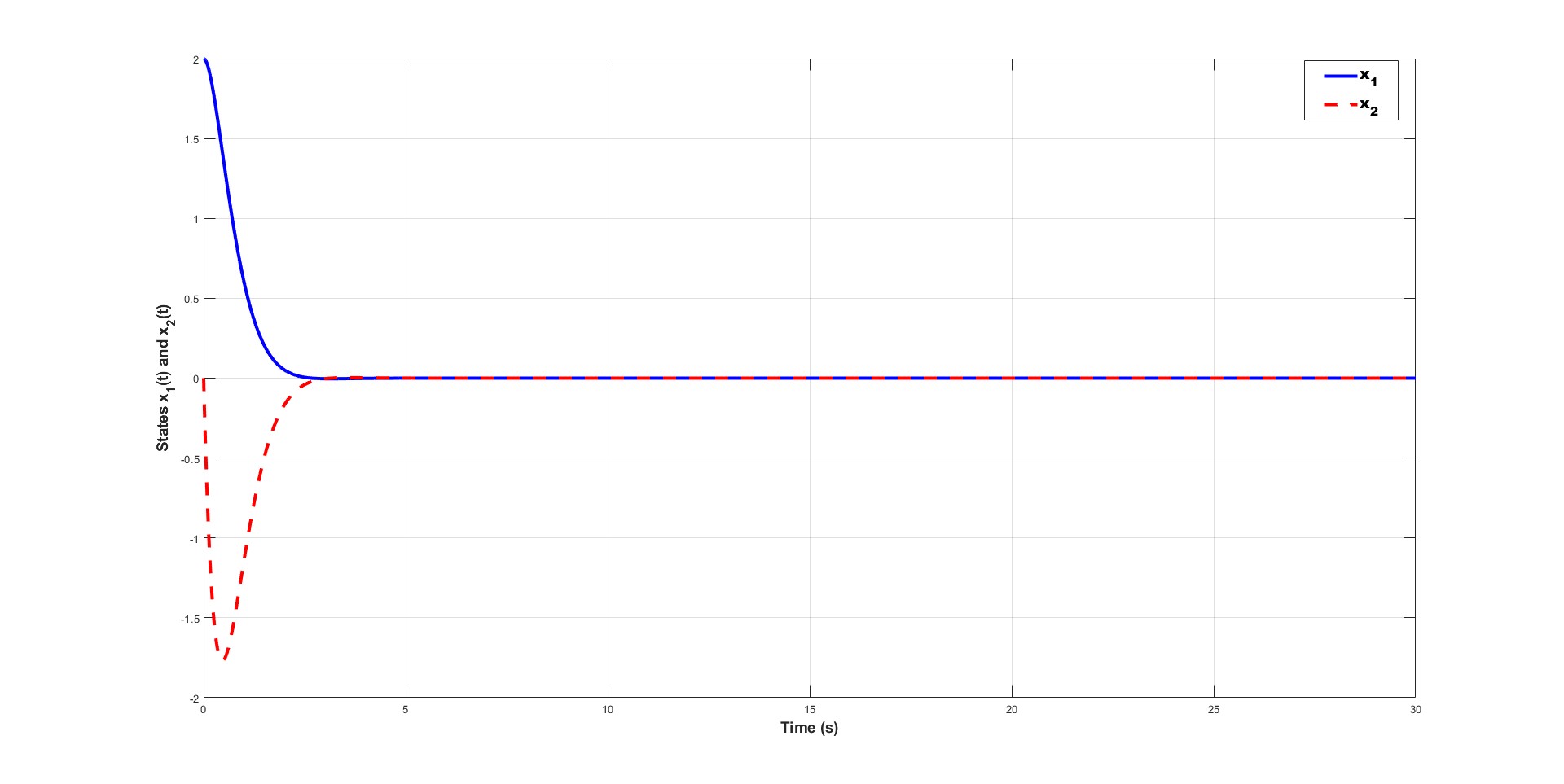}
    \caption{State stabilization of the Van der Pol oscillator with initial condition $\mathrm{x}_{0}=[2\, \, 0]$.}
    \label{fig:fdstateslinearstabilized}
\end{figure}

\begin{figure}[H]
    \centering
    \includegraphics[width=1.00\linewidth]{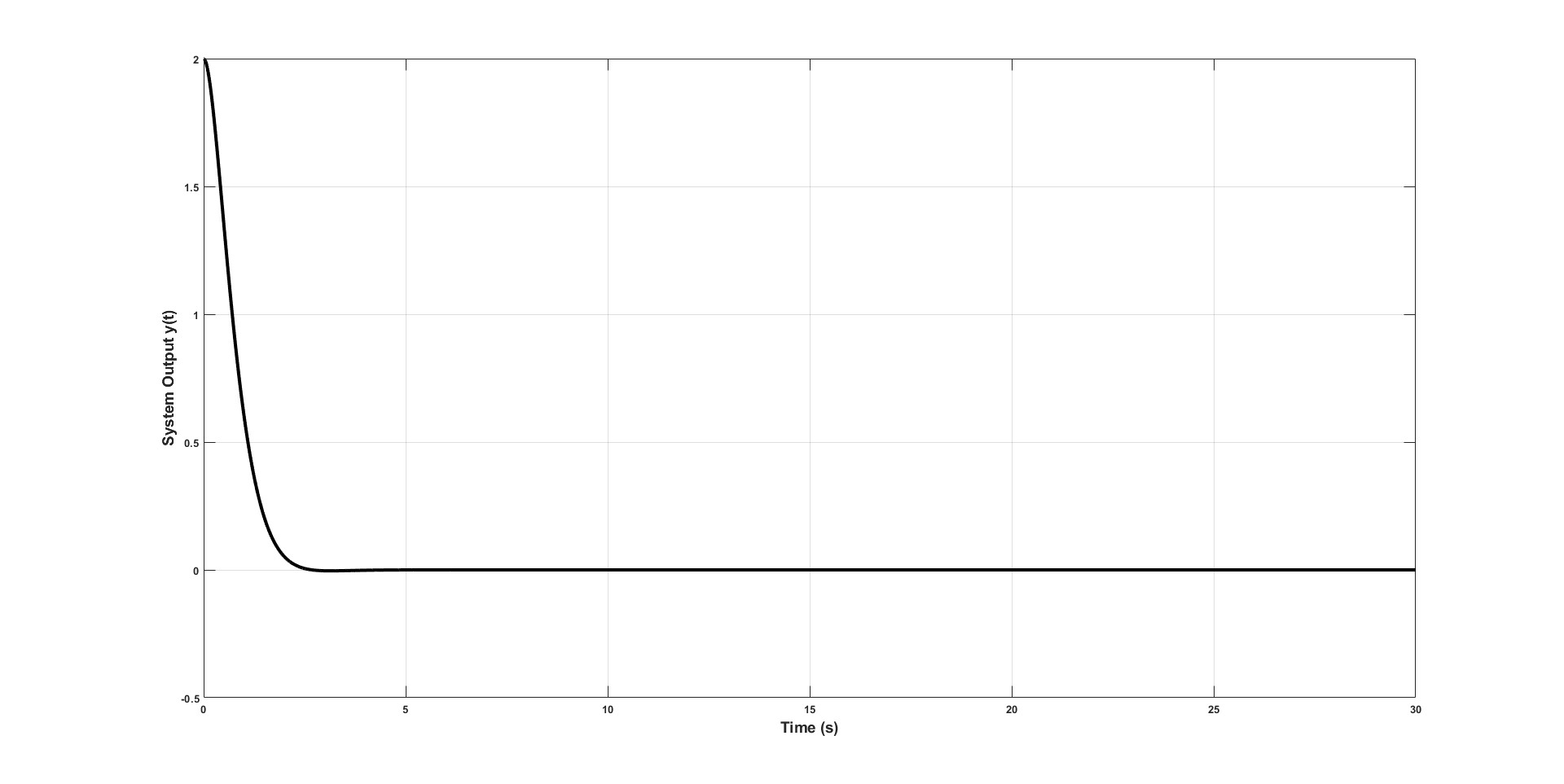}
    \caption{Output stabilization of Van der Pol oscillator.}
    \label{fig: fdoutput}
\end{figure}

  \begin{figure}[H]
    \centering
    \includegraphics[width=1.00\linewidth]{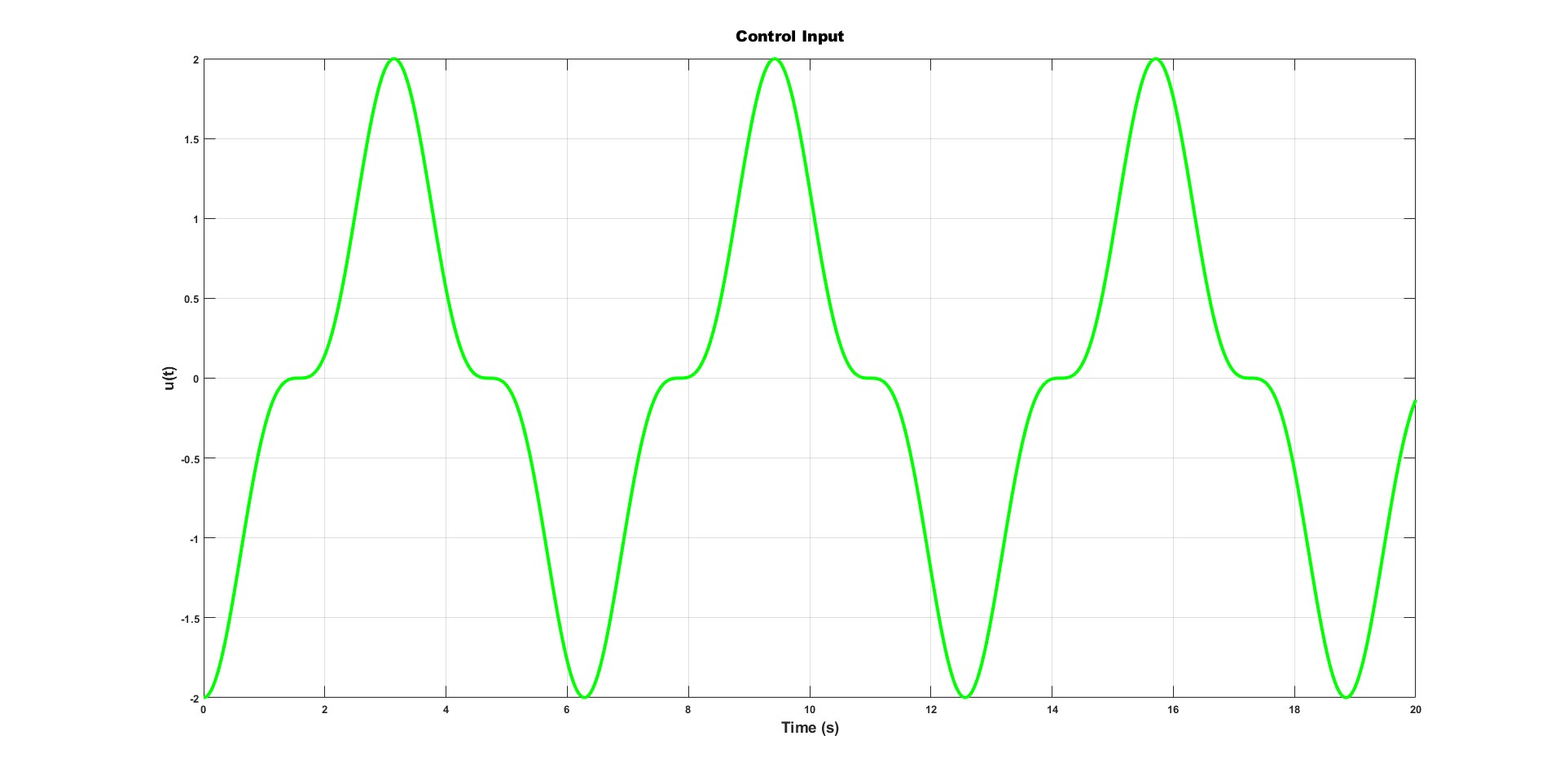}
    \caption{Tracking controller with reference signal $r(\mathpzc{t})=\sin(\mathpzc{t})$.}
    \label{fig: utref}
\end{figure}
  
\begin{figure}[H]
    \centering
    \includegraphics[width=1.00\linewidth]{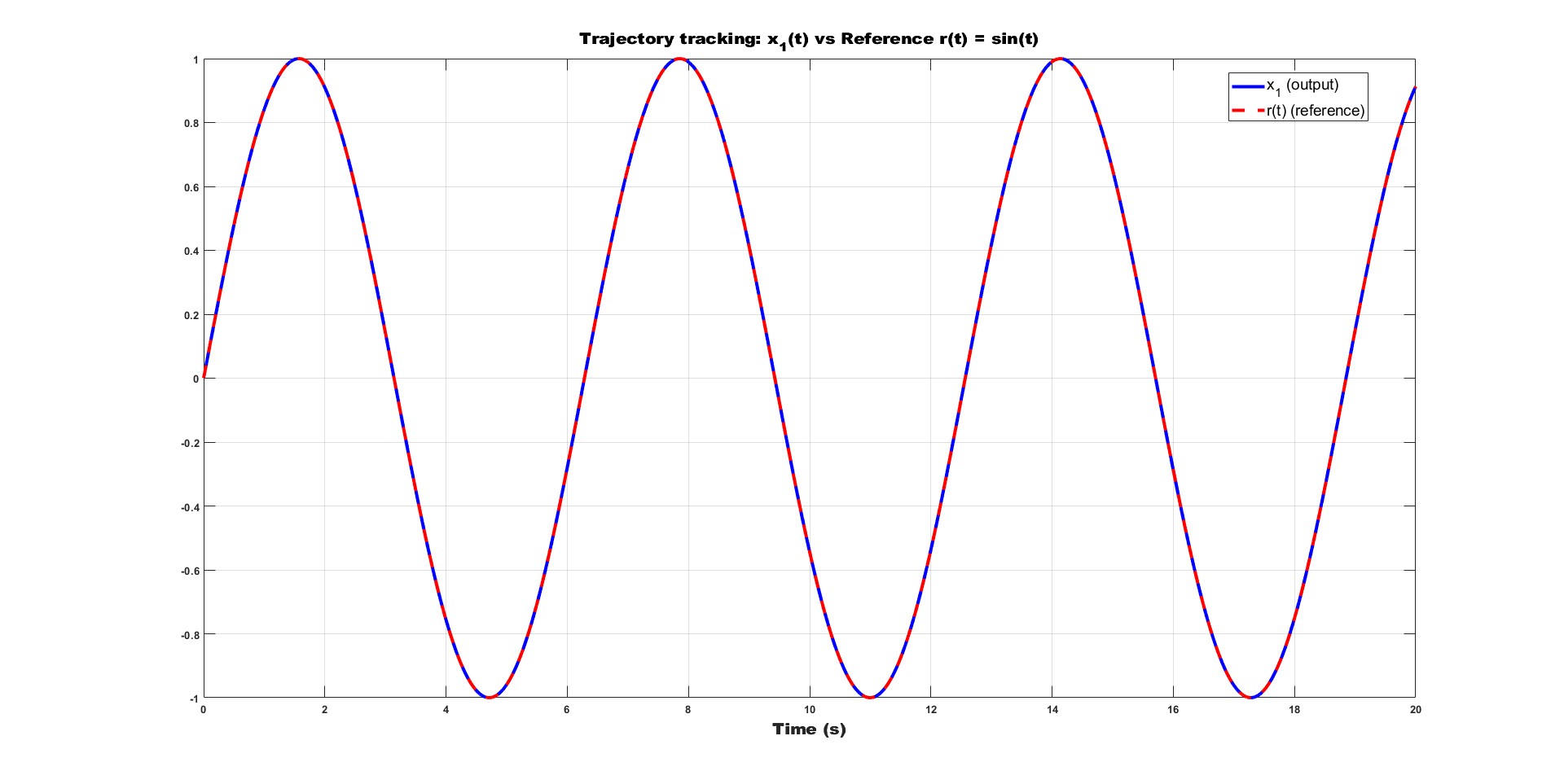}
    \caption{Output trajectory tracking of feedback linearized Van der Pol oscillator.}
    \label{fig: reftrack}
\end{figure} 

\begin{figure}[H]
    \centering
    \includegraphics[width=1.00\linewidth]{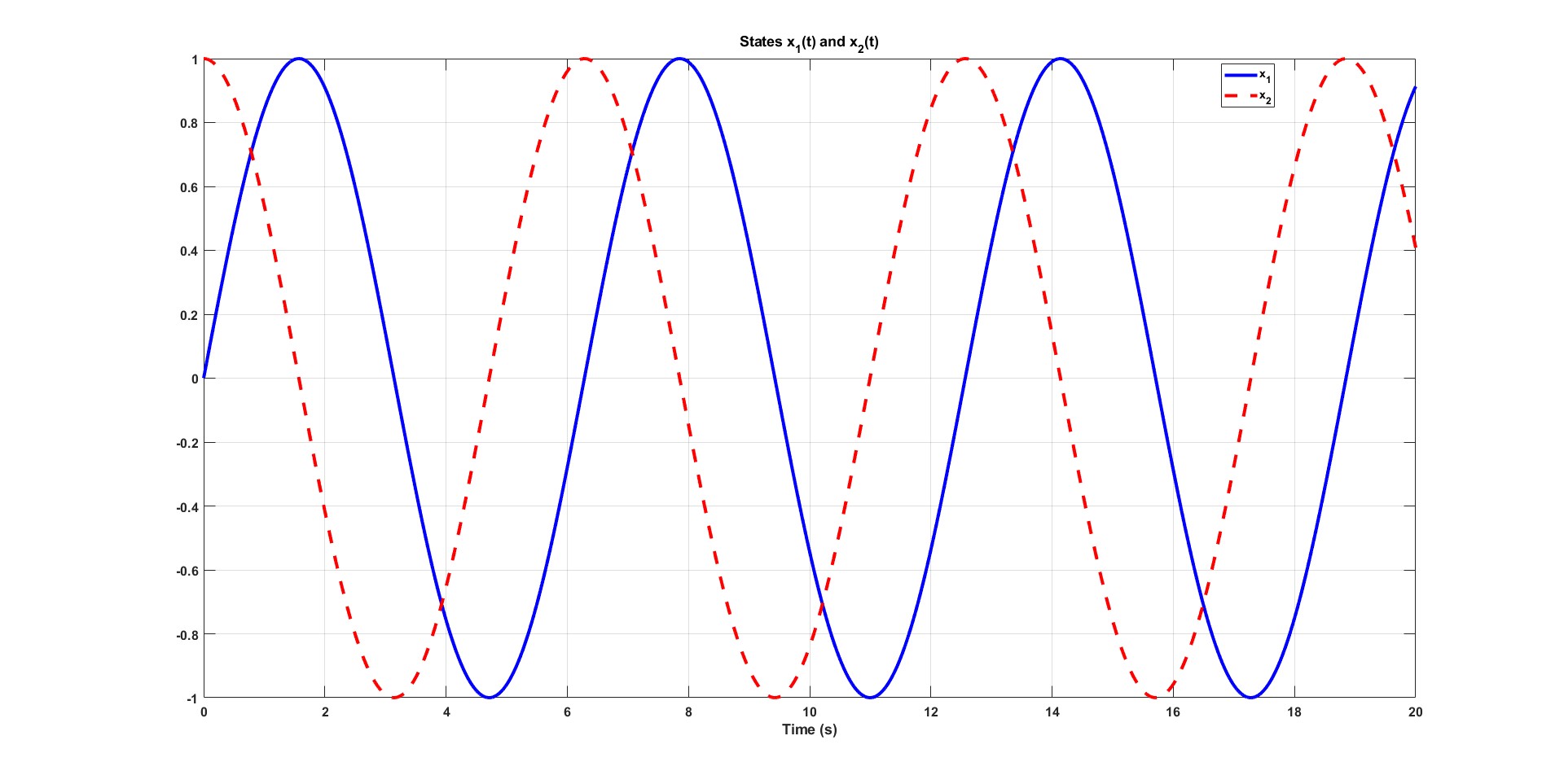}
    \caption{Trajectory tracking of feedback linearized controller with $\mathrm{x}_{d}=[\sin\mathpzc{t}, \cos\mathpzc{t}]$. }
    \label{fig: refstates}
\end{figure} 
The above graphical visualizations from figure 1-9, clearly demonstrates that the proposed stacked regression based identification framework, together with the derived bilinear constraint, evidently feedback linearized and stabilized the two-state Van der Pol oscillator. Notably, this is accomplished without requiring any explicit prior knowledge of the original nonlinear system equations.

\section{Conclusion}
This study introduces a novel approach of stacked regression to identify the true governing equations of a nonlinear physical control system. The proposed methodology begins by constructing separate libraries for the system input and output using the available datasets. A stacked sparse regression technique is then employed to formulate a minimization problem, augmented with a constraint designed to enforce feedback linearization of the identified system.
This constraint is derived using the concepts of Lie derivatives and the relative degree condition. A threshold parameter, $\uplambda$ , is introduced to promote sparsity in the regression. The validity of the results is demonstrated through mathematical induction. Initially, the case where the relative degree $r=2$
corresponding to system with a two dimensional state vectors is analyzed. The findings are then generalized and proven for system with an arbitrary relative degree $r=n.$ The effectiveness of the proposed approach is substantiated through a relevant case study.

\end{document}